\documentclass[twocolumn,aps]{revtex4-2}

\usepackage{graphicx}   
\usepackage{natbib}     
\usepackage{hyperref}   
\usepackage{amsmath,amssymb} 
\usepackage{newtxtext,newtxmath} 

\begin{document}

\title{Gaussian Process Surrogate Models for Efficient Estimation of Structural Response Distributions and Order Statistics}

\author{Vegard Flovik}
\email{vegard.flovik@dnv.com}
\author{Sebastian Winter}
\author{Christian Agrell}
\affiliation{Group Research and Development, DNV, Norway}

\begin{abstract}
Engineering disciplines often rely on extensive simulations to ensure that structures are designed to withstand harsh conditions while avoiding over-engineering for unlikely scenarios. Assessments such as Serviceability Limit State (SLS) involve evaluating weather events, including estimating loads not expected to be exceeded more than a specified number of times (e.g., 100) throughout the structure's design lifetime. Although physics-based simulations provide robust and detailed insights, they are computationally expensive, making it challenging to generate statistically valid representations of a wide range of weather conditions.

To address these challenges, we propose an approach using Gaussian Process (GP) surrogate models trained on a limited set of simulation outputs to directly generate the structural response distribution. We apply this method to an SLS assessment for estimating the order statistics \(Y_{100}\), representing the 100th highest response, of a structure exposed to 25 years of historical weather observations. Our results indicate that the GP surrogate models provide comparable results to full simulations but at a fraction of the computational cost.
\end{abstract}

\maketitle

\section{Introduction}\label{intro}
Accurate estimation of structural responses under diverse weather conditions is influenced by both the variability of the weather environment (e.g., waves, wind, currents) and the variability of the structural response in a given random weather state. For precise long-term estimation, it is essential to consider both these variabilities. 

Order statistics, which involve analyzing specific ranked values within a dataset, are particularly useful in this context. These statistics can involve extreme values like the maximum or the minimum,  as well as other values such as the 100th largest response. By examining these ranked values, order statistics provide valuable insights into the behavior of structures under various conditions, which is crucial for both reliability and serviceability assessments. 

While traditional physics-based simulation methods can calculate order statistics, this is often impractical due to the computational expense. This is especially true when dealing with long time periods, such as 25 or 100 years, which are typically used in the design of structures \citep{wang2022reliability}.

A common practice in the engineering field is thus to use surrogate models, which approximate the results of high-fidelity simulations. 
These models, such as Gaussian Process (GP) models, can achieve similar accuracy with significantly reduced computational cost \citep{samadian2024application}.

In this paper, we propose a method for creating GP-based surrogate models suitable for order statistics calculation.  Our approach assumes that the structural characteristics remain constant during the studied timeframe, which is a common simplification in practical structural response simulations \citep{DNV-RP-C205}.

Our method introduces several aspects that distinguish it from existing methods. Specifically, we do not use surrogate models to estimate the structural responses directly. Instead, we estimate the parameters of the structural response distribution. This allows us to generate samples from the predicted distribution, enabling efficient calculation of order statistics without the need for extensive simulations. Additionally, our approach is designed to work with stochastic simulators where both the responses and the number of data points returned vary stochastically. 

Our method is particularly valuable for Serviceability Limit State (SLS) calculations \citep{DNV-ST-0119}, where the evaluation of structural responses under a wide range of weather conditions is crucial but difficult to achieve with traditional methods like environmental contours \citep{vanem2020comparing}.

To demonstrate our method, we conducted an SLS assessment estimating the 100th largest response ($Y_{100}$) for a structure exposed to 25-years worth of historical weather observations. This proof-of-concept uses a simplified stochastic simulation model that balances realistic dynamics and computational efficiency. We benchmark our method against a brute-force approach that calculates order statistics directly using the simulator. Our method showed comparable results at a fraction of the computational cost.

\section{Problem Statement and Approach}\label{problem_statement}

The specific problem addressed in this paper is the need for an efficient and accurate method to estimate the order statistics, \( Y_k \), representing the \( k^{\text{th}} \) largest response within a selected time interval. For systems where the response is stochastic, this is challenging using traditional methods due to the inherent variability of the responses, which would require a high number of simulations to capture accurately.

Our proposed method maps weather data inputs to predicted distributions of structural responses using a surrogate model, and then generates data to mimic the simulator output. This enables efficient estimation of selected order statistics, effectively bypassing the need for generating the structural responses using a simulator.

Our approach is inspired by \cite{preprint}, which uses a Gaussian Process to model the parameters of the output distribution. Here, we extend this method by not only modeling the distributional parameters, but also generating realizations of the predicted structural response from the predictive distributions.

The simulator is considered to be a stochastic black-box function, represented by
\begin{equation}
\mathrm{sim}(\mathbf{x}) \to [R_1, \dots, R_{L|\mathbf{x}}],
\end{equation}

where each \( R_i \) represents the response within a certain time interval, as explained in further detail in Section \ref{model_intro}. The number of values returned by the simulator, \( L|\mathbf{x} \), is a random variable conditional on \( \mathbf{x} \). This means that both the responses \( R_i \) and the count \( L \) are stochastic outputs of the simulator.

We assume outputs of the simulator at a point \( \mathbf{x} \) are samples from a distribution \( R|\theta_R(\mathbf{x}) \), governed by the underlying physics of the system, where \( \theta_R(\mathbf{x}) \) are the parameters of the distribution. For example, if \( R \) is a Gumbel distribution, then \( \theta_R(\mathbf{x}) = (\textit{location}(\mathbf{x}), \textit{scale}(\mathbf{x})) \). In other words, we assume a fixed distribution type with unique parametrization at each \( \mathbf{x} \).

Producing the surrogate model's estimate of a single \( \mathrm{sim}(\mathbf{x}) \) run works as follows:
\begin{enumerate}
    \item Use the Gaussian Process to map \( \mathbf{x} \to \theta_R, L \).
    \item Create the distribution \( R \) using the parameters \( \theta_R \).
    \item Generate a sample from \( L|\mathbf{x} \), then generate \( L \) samples from distribution \( R \), representing the output of the simulation model.
\end{enumerate}

While this mapping could be performed with many different models, using a Gaussian Process allows us to quantify the uncertainty in our estimation, as well as propagating uncertainty about the true surrogate model to our estimates of the order statistics \( Y_k \).

The GP model assumes that the function mapping inputs to outputs is a realization of a Gaussian process, defined by its mean function and covariance function, which encodes certain assumptions about the function we aim to predict, such as smoothness properties or periodicity. In this study, we use the Matérn covariance function, which is suitable for modeling functions with varying smoothness \citep{gaussianprocesses}.

Further details on the simulation model, surrogate model, and quantities of interest calculation are covered in the following sections.

\section{Methods}
This section details the application of our approach to a proof-of-concept use-case.

\subsection{Simulation model overview}
The simulator used in this study takes weather observations as input, specifically significant wave height, \(H_s\), peak wave period \(T_p\), and wind speed \(V_w\), to calculate the structural response of a marine structure. 
We use real weather observations collected from 1979 to 2015, spanning slightly over 36 years. Each observation represents the average conditions experienced over one hour.

The simulation involves several key steps, which are outlined below for completeness. However, note that the method presented in this paper does not utilize any of these internal details and treats the simulator as a stochastic black-box function.

\subsubsection{From weather data to structural responses and simulator output}\label{model_intro}
From the observed weather data, a wave spectrum is generated using the Torsethaugen model \citep{torsethaugen1996model}. Since the weather data represent average values measured over one hour, multiple wave spectrum realizations can satisfy these inputs, introducing stochasticity into the simulation model. 
The response spectrum of the structure is then calculated by combining this wave spectrum with the structure's transfer function \citep{bendat2011random}. The response spectrum is then transformed into a time-domain series using an inverse Fourier transform \citep{oppenheim_signalprocessing}. The wind-induced moment is calculated using the wind speed and a predefined thrust curve. It is then combined with the wave-induced bending moments to determine the response of the structure.

\begin{figure}[t]
    \centering
    \includegraphics[width=.95\linewidth]{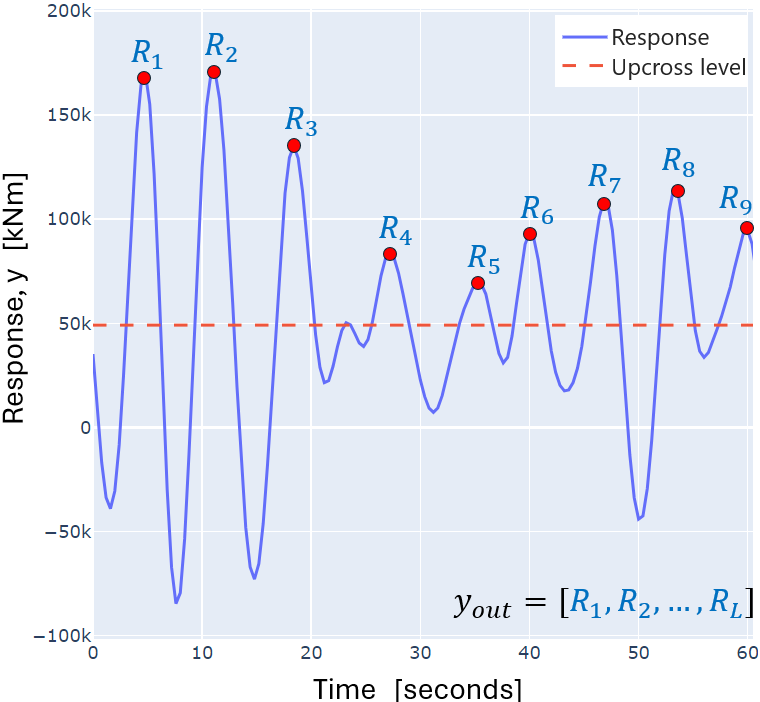}
    \caption{Example output from simulation model: array of $L$ peak values, $y_{out}$}
    \label{fig:simulator_output}
\end{figure}

The time series of the structural response is then analyzed to identify up-crossings and peak values \citep{ochi1998ocean}. As illustrated in Figure \ref{fig:simulator_output}, up-crossings occur when the response exceeds a specified threshold (the mean response over the time segment in this case), and the highest response (peak value) within each up-crossing interval is recorded. The final output of the simulation model is then an array of \(L\) peak responses.

\subsection{Calculating order statistics}\label{sec:method:qoi}
Our Quantity of Interest (Q.O.I) is the distribution of the 100th highest response, \(Y_{100}\), across the selected time interval, here 25 years. 
Each combination of input variables, \(\{H_s, T_p, V_w\}\), corresponds to one hour of observed weather data. Running the simulator on an input results in an array of $L$ peak responses, as explained in the previous section.

To produce a brute force calculation of $Y_{100}$, we first create an empty list ($\textit{top100}$) to store the overall 100 largest responses seen in the 25 year period. We then run the simulator on the 25 years worth of input samples. Each time the simulator runs we update $\textit{top100}$ if any of the responses created in that run are large enough to warrant inclusion. Once all the data has been run, the 100th largest value in $\textit{top100}$ is extracted to provide a realization of $Y_{100}$.
\begin{figure}[t]
    \centering
    \includegraphics[width=1.\linewidth]{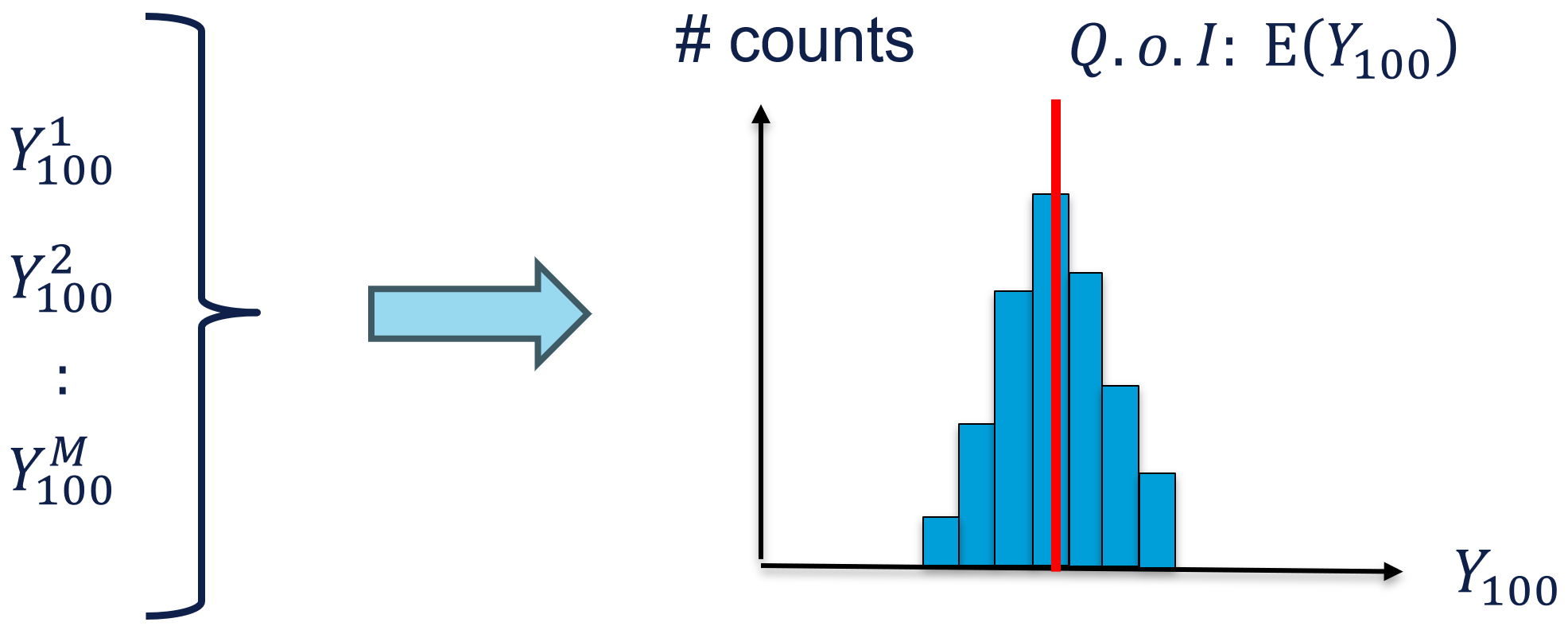}
    \caption{Estimating the expectation value of $Y_{100}$ as the Q.O.I}
    \label{fig:QOI_Estimate}
\end{figure}

Given that each data point represents one hour, the simulation count becomes \(N = 24 \times 365 \times 25 = 219.000\). 
Since the simulator output is stochastic, and \(Y_{100}\) is a derived quantity from this process, we perform multiple realizations (\(M = 100\)) to estimate the distribution of possible values, as illustrated in Figure \ref{fig:QOI_Estimate}.
For a time period of 25 years, this results in \(N \times M \approx 22\) million simulation runs. Our simplified simulator is specifically designed for this to be feasible, but real-world simulators are typically far too slow or expensive to run this many times.

\subsection{Generating training data for the surrogate model}

To generate training data for the surrogate model, we first define the input variable ranges based on the historical weather observations. We then use uniform random sampling to draw \(N=5000\) points within this range, and for each set of input variables, we generate \(M\) samples from the simulation model to capture its stochastic nature. The relatively high number of data points represents a best case scenario for training the GP models. For a more realistic setup with a computationally expensive simulation model, we could make use of Design of Experiments (also referred to as active learning) to iteratively build a dataset in a more efficient way \citep{moustapha2022active}.

\begin{figure}[t]
    \centering
    \includegraphics[width=.95\linewidth]{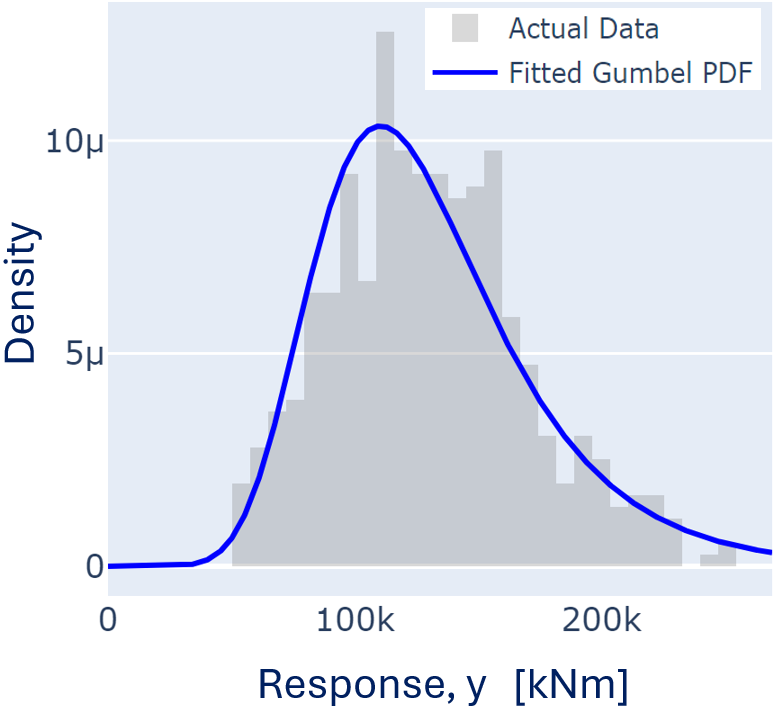}
    \caption{Fitting the output from the simulation model to a Gumbel distribution.}
    \label{fig:response_distr}
\end{figure}

For each data point we fit a distribution to each of the $M$ runs, as illustrated in Figure \ref{fig:response_distr}. We then have $M$ estimates of the distribution parameters for that point. We calculate the mean and standard deviation over the $M$ parameter estimates. We repeat this process for three different distributions: Gumbel, Rayleigh, and Weibull. The standard deviation quantifies the noise level, or variability, of the generated data due to the stochasticity of the simulations. The resulting dataset, as illustrated in Table \ref{training_data}, is then split into a training set (80\%) and a test set (20\%) used to evaluate the model predictions

\begin{table}[b]
\caption{Generated dataset for training the GP models, where the weather observations \(\{H_s, T_p, V_w\}\) represent the input features and the distributional parameters represent the target variables.}
\label{training_data}
\centering
\begin{tabular*}{\columnwidth}{@{\extracolsep{\fill}} r r r r r r r r @{}} 
\toprule 
\( H_s \) & \( T_p \) & \( V_w \) & \( \mu \) & \( \sigma_{\mu} \) & \( \beta \) & \( \sigma_{\beta} \) & \dots\\ 
\colrule
3.2 & 11.3 & 1.2 & 75371 & 891 & 20983 & 530 & \dots\\
7.4 & 9.6 & 16.5 & 236624 & 1505 & 46570 & 1066 & \dots\\
\vdots & \vdots & \vdots & \vdots & \vdots & \vdots & \vdots & \dots\\
\botrule
\end{tabular*}
\end{table}

The choice of these distributions is justified by their relevance and applicability to modeling environmental and structural response data. The Gumbel distribution is commonly used to model the distribution of extreme values, \citep{coles2001introduction} which may make it suitable for modeling the structural responses that may occur under severe weather conditions.  
The Rayleigh distribution is often used to model wave heights and wind speeds in oceanographic and meteorological studies \citep{liu2012extreme}. It may thus be appropriate for representing the distribution of responses that are influenced by the combined effect of multiple independent factors, which is characteristic of wave and wind data. 
The Weibull distribution is versatile and can model a wide range of data types, including weather data \citep{tucker2001waves}. Its flexibility may make it suitable for capturing the variability in structural responses under diverse weather conditions. 

Testing a selection of different distributions is essential because it is not straightforward to decide a priori which distribution best maps the data. This also evaluates the sensitivity of our method to the choice of distribution, which is harder to assess in more complex cases where the true output from the simulator is not available for direct comparison.

\begin{figure*}[t]
    \centering
    \includegraphics[width=1.\linewidth]{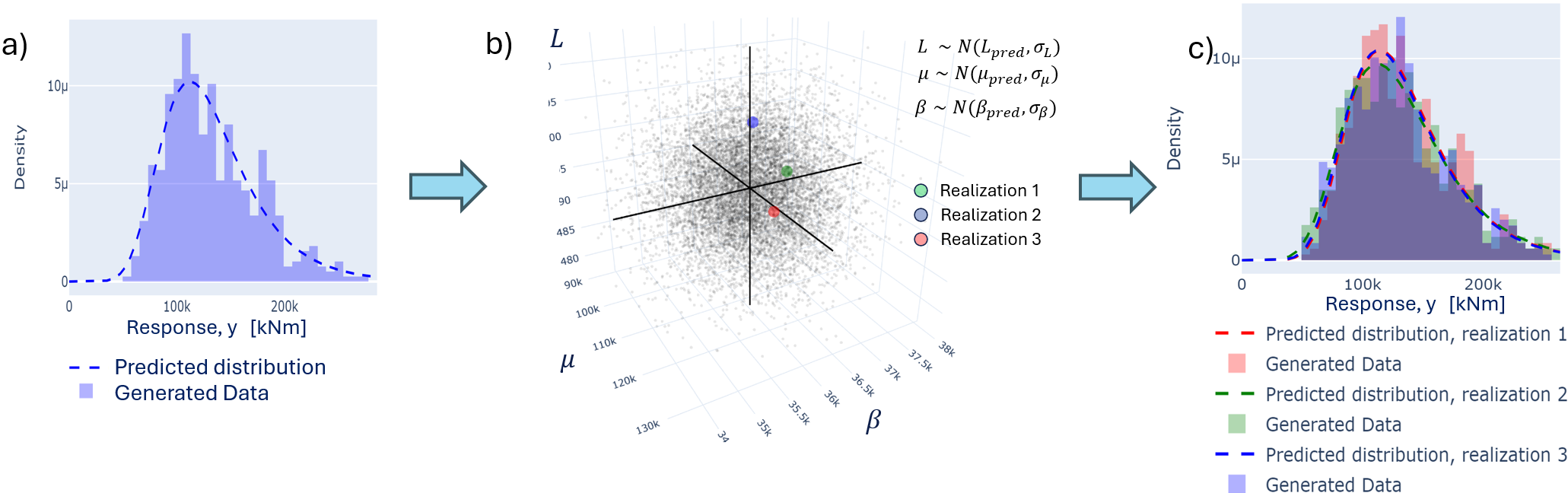}
    \caption{From point predictions to multiple realizations. a) Single prediction and the corresponding generated data. b) Sampling multiple realizations from the GP's predictive distribution. c) Generating data using multiple realizations to estimate the uncertainty in the predicted responses.}
    \label{fig:sampling_responses_2}
\end{figure*}

\subsection{Generating the structural response spectrum using GP models}

\subsubsection{Training the models}
Based on the dataset from the previous section, we train a series of GP models to predict the distribution parameters directly from the weather observations. When training the GP models, the standard deviations are used to quantify the noise level in the training data, capturing the inherent stochasticity of the simulator output. We use a separate model for each target variable. For example, Model 1 predicts \((\mu, \sigma_\mu)\), Model 2 predicts \((\beta, \sigma_\beta)\), etc.
\begin{equation}
\mu, \sigma_\mu = \text{GP}_\mu(\{H_s, T_p, V_w\})
\end{equation}

\subsubsection{Generating the predicted response distribution}

\begin{figure}[t]
    \centering
    \includegraphics[width=.9\linewidth]{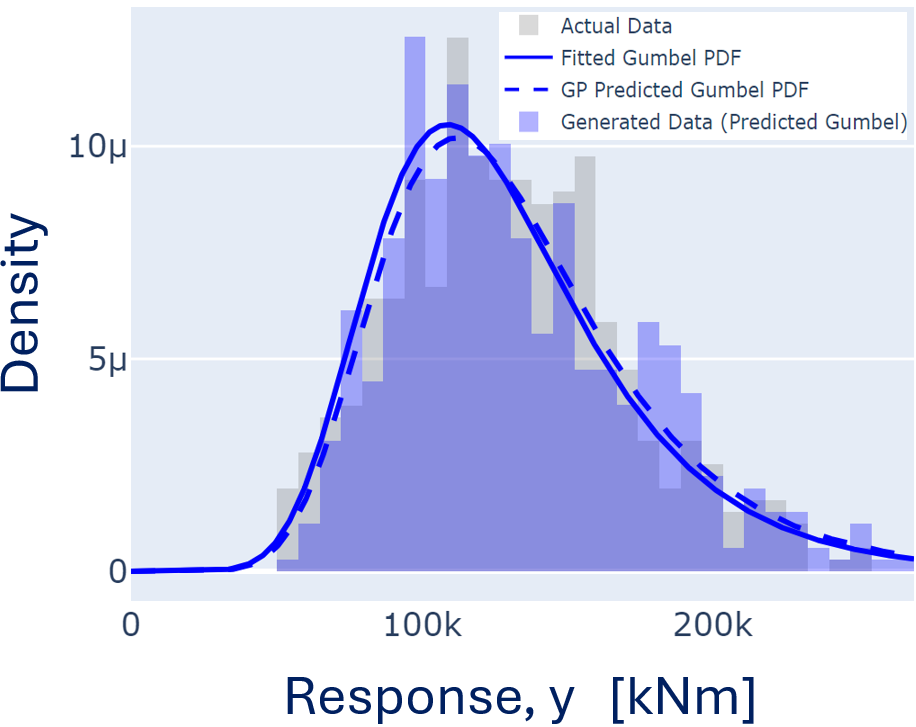}
    \caption{Generating structural response data from predicted distribution.}
    \label{fig:response_distribution}
\end{figure}

For each combination of \(\{H_s, T_p, V_w\}\), we first predict the distribution parameters, e.g. \(\mu \) and \(\sigma_\mu\), and  the length \(L\) of the response vector. We then generate \(L\) data points from this distribution, as illustrated in Figure \ref{fig:response_distribution}. 

A key advantage of using GPs as surrogate models is their ability to estimate the uncertainty of their predictions. This uncertainty can then be propagated to assess its impact on the generated responses.
Figure \ref{fig:sampling_responses_2} illustrates this process.

Each time the surrogate is run we sample a set of parameters from the GP's predictive distribution. These sampled parameters are then used to generate the structural responses. This method effectively transfers uncertainty from the GP predictions to the generated output data, capturing the variability in the structural response.

\section{Results and Discussion}\label{Results}
\begin{figure}[]
    \centering
    \includegraphics[width=1.\linewidth]{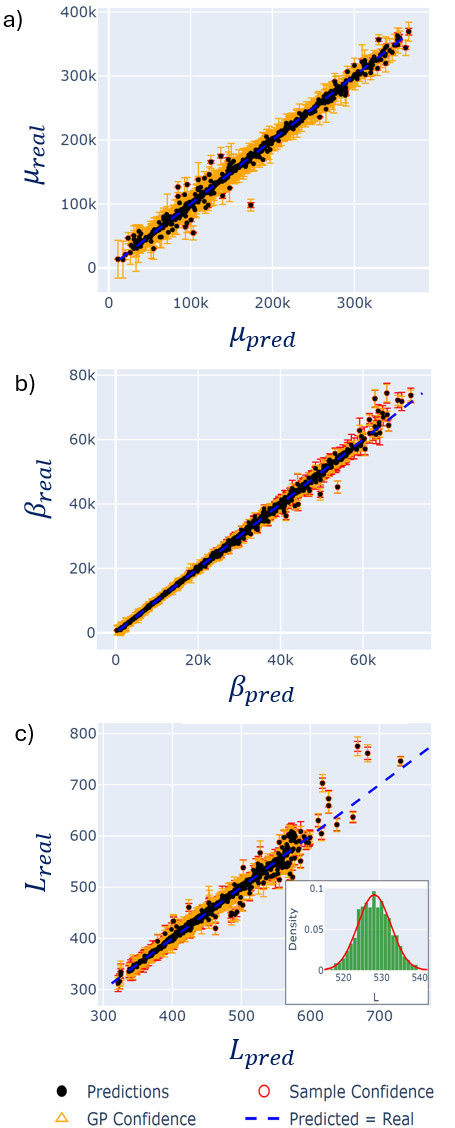}
    \caption{Examples of predicted parameters for the Gumbel distribution, validated on hold-out test set. a) Real. vs predicted \(\mu \). b) Real vs. predicted \(\beta \). c) Real vs. predicted \(L \). Inset: illustrating distribution of length parameter.}
    \label{fig:test-set_evaluation}
\end{figure}

For the partitioned test set, we predicted the distributional parameters and compared them with the true values to check the quality of fit. An example of the predicted parameters for the Gumbel distribution is illustrated in Figure \ref{fig:test-set_evaluation}, where the error bars indicate the standard deviation. 
The length of the response vector, \(L \), can be predicted directly, as it follows a Gaussian distribution (see inset for sampled distribution of \(L \) in Figure \ref{fig:test-set_evaluation}c).

We estimate the Q.O.I using the surrogate model by running the same brute force calculation process described in Section \ref{sec:method:qoi}, but making predictions with the surrogate rather than the simulator. Our results showed that responses generated from the Rayleigh and Weibull distributions closely matched the true output from the simulator. In contrast, the Gumbel distribution tends to overestimate the larger responses due to its  heavier tails. The Rayleigh and Weibull distributions thus appear better suited for modeling the responses from this simulator because of their lighter tails and greater flexibility in fitting a wide range of data.

\subsection{Estimation of \(Y_{100}\) over a 25 year period}

The dataset spanning 25 years of historical weather observations covers a wide range of weather conditions, providing a solid basis for evaluation.
Based on the results from the previous section, we restricted our evaluation to data generated using the Weibull and Rayleigh distributions, as the Gumbel distribution showed unsatisfactory performance.

\begin{figure}[b]
    \centering
    \includegraphics[width=1.\linewidth]{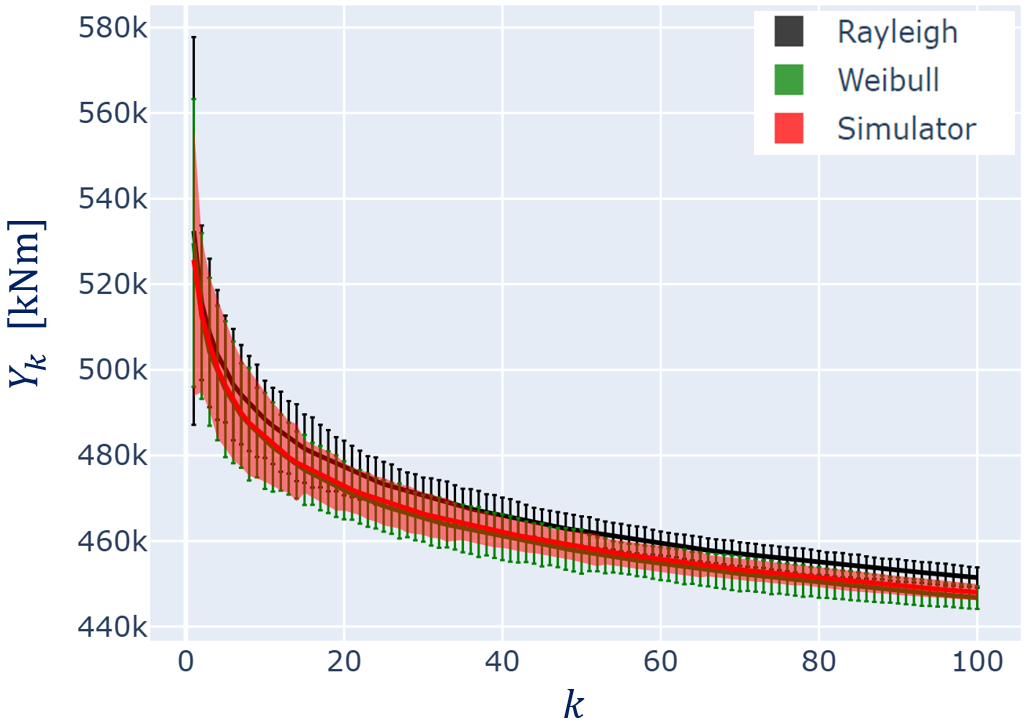}
    \caption{Estimated values for the $k^{th}$ largest responses, $Y_{k}$, with 95\% confidence intervals for both GP predictions (shown as error bars) and simulator output (represented by the shaded region).}
    \label{fig:pred_vs_real_25year}
\end{figure}

We iterated through all observations of \(\{H_s, T_p, V_w\}\) and sampled \(M=100\) realizations from the GP's predictive distribution. 
We then compared the generated output from the GP models with the responses computed directly using the simulation model. Also here, we performed \(M=100\) realizations for each set of weather observations to account for the inherent stochastic nature of the simulations. 
These results are illustrated in Figure \ref{fig:pred_vs_real_25year}, where we compare the top 100 responses, \(Y_k\), for the Rayleigh, Weibull, and simulator outputs.

\begin{figure}[t]
    \centering
    \includegraphics[width=1.\linewidth]{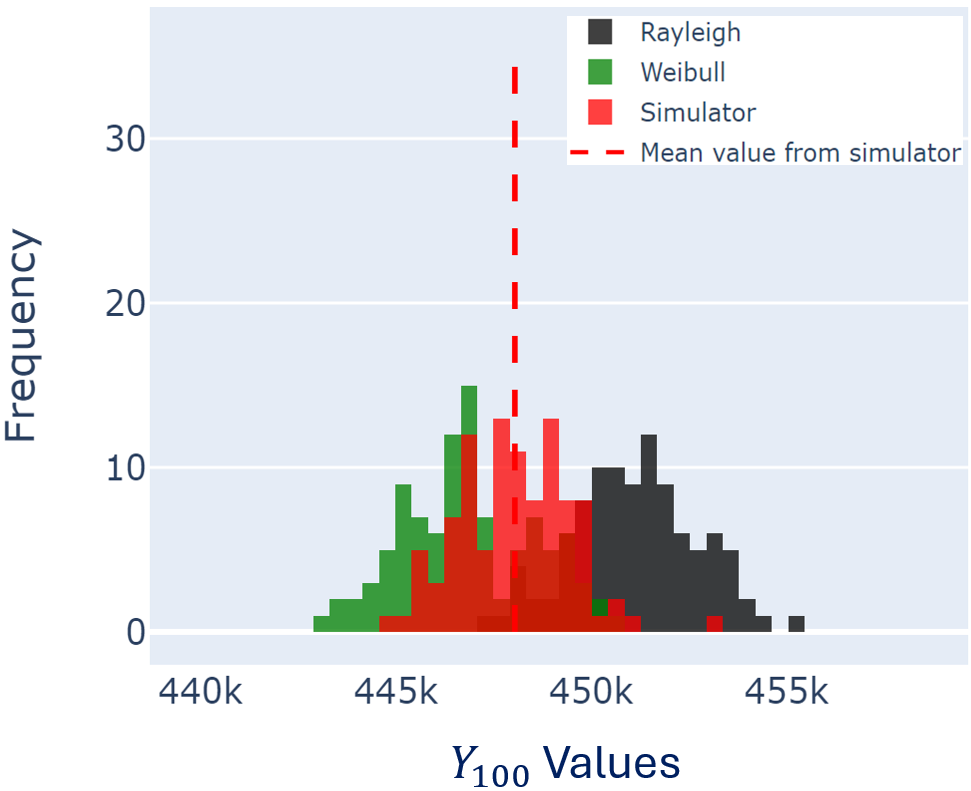}
    \caption{Histogram of $Y_{100}$, as estimated from GPs vs. simulation model. }
    \label{fig:Y_100_25year}
\end{figure}

In our case, we are explicitly interested in the 100th highest response i.e., \(Y_{100}\). 
To examine this quantity in greater detail, we plot a histogram of \(Y_{100}\), comparing the GP predictions with the simulator output.

These results are illustrated in Figure \ref{fig:Y_100_25year}, indicating that the structural responses generated from the GP models align well with the output from the simulation model. The Weibull results match the true distribution more closely, but provide a non-conservative estimate. In contrast the Rayleigh results are conservative, producing results similar to \(Y_{80}\) estimated by the simulator. The difference in results is not sufficiently large to draw definitive conclusions about the suitability of the distributions. In practice, when using this method without the true simulator results available for direct comparison, engineers could apply a variety of reasonable distributions to the problem and select one of the more conservative Q.O.I estimates. 

The choice of distribution is an influential decision in this method, and further work should be undertaken to provide guidelines for selecting appropriate distributions a priori.

\section{Conclusions}\label{Conclusions}

In summary, our study demonstrates that GP surrogate models offer a promising approach for efficient and accurate estimation of order statistics. This method is especially beneficial for calculating key order statistics and performing SLS assessments in stochastic systems, where traditional simulation methods are impractical. 

Our proof-of-concept use-case was based on a simplified simulation model. However, given the positive results, we believe this method is worth exploring further. Future work should focus on applying this method to more complex simulation models and real-world datasets to validate its effectiveness and generalizability. Additionally, exploring the integration of Design of Experiments (DOE) techniques to optimize the training data generation process could further enhance the efficiency and accuracy of the GP models.

If proven successful, this method would allow for a more detailed statistical representation of weather-induced structural responses, leading to better-informed design decisions for structures exposed to varying weather conditions.

\section*{Acknowledgement}
This research was conducted as part of the Norwegian Research Council funded project RaPiD (Reciprocal Physics-based and Data-driven models, grant no. 313909).

\bibliography{References}
\end{document}